\documentclass[letterpaper, 10 pt, conference]{ieeeconf}  

\IEEEoverridecommandlockouts                              

\usepackage{cite}
\usepackage{amsmath,amssymb,amsfonts}
\usepackage{graphicx}
\usepackage{textcomp}
\usepackage{xcolor}
\usepackage{orcidlink}
\usepackage{times}
\usepackage{epsfig}
\usepackage{amssymb}
\usepackage{flexisym}
\usepackage{booktabs}
\usepackage{multirow}
\usepackage{array}
\usepackage{tabularx}
\usepackage{ragged2e}
\usepackage{multicol}
\usepackage{subfigure}
\usepackage{gensymb}
\usepackage{tabstackengine}
\usepackage{adjustbox}
\usepackage{etoolbox}
\usepackage{bm}
\usepackage{threeparttable}

\usepackage[all]{nowidow}
\usepackage[ruled,vlined]{algorithm2e}

\title{\LARGE \bf
FLEX-WAM: Flexible Block-Causal World–Action Models for Long-Horizon Imagination and Planning
}

\author{R. Khorrambakht$^{*1}$, Joseph Amigo$^{*1}$, Félix Lebel$^{1}$, Leon Seetoo$^{1}$,\\ Jean Ponce$^{2,3}$, Zhenzhen Li$^{4}$, and Ludovic Righetti$^{1,5}$
\thanks{$^{*}$Equal contribution / These authors contributed equally to this work.}%
\thanks{$^{1}$ Center for Robotics and Embodied Intelligence (CREO), New York University}
\thanks{$^{2}$ Courant Institute of Mathematical Sciences and Center for Data Science, New York University}
\thanks{$^{3}$ Ecole normale supérieure - PSL}
\thanks{$^{4}$ NVIDIA}
\thanks{$^{5}$ Artificial and Natural Intelligence Toulouse Institute (ANITI)}
}

\DeclareMathAlphabet\mathbfcal{OMS}{cmsy}{b}{n}
\mathcode`*=\string"8000
\begingroup
\catcode`*=\active
\xdef*{\noexpand\textup{\string*}}
\endgroup

\begin{document}

\maketitle

\begin{abstract}

World--action models (WAMs) promise a unified model that predicts action-conditioned futures, generates feasible actions, and supports planning in imagination. However, existing joint video--action models often use computationally heavy, fixed-horizon backbones ill-suited to streaming inference and stable long-horizon open-loop rollouts. We introduce FLEX-WAM, a Flexible and Efficient Block-Causal World--Action Model for unified simulation and policy inference. FLEX-WAM supports variable-length contexts and non-causal prediction horizons, as well as infinite autoregressive generation frame by frame or block by block. Its block-causal, KV-cacheable architecture combines axial attention and blockwise diffusion forcing to enable efficient real-time rollout and deployment-time latency--throughput tradeoffs without retraining. Joint training can nevertheless produce plausible futures that weakly respond to commanded actions. We address this failure mode by balancing state and action flow-matching gradient contributions across the state--action diffusion-noise grid and regulating world-model sampling using \emph{Forward-Dynamics (FD) elasticity}, an efficient training-time proxy for action responsiveness. Across simulated and real-world datasets, FLEX-WAM achieves superior multi-step prediction quality and latency while producing stable joint state--action rollouts for thousands of steps. As a joint action proposer and simulator within MCTS, it solves long-horizon PushT and all five OGBench Puzzle-4x4 tasks entirely in imagination. On a bimanual OpenArm-based robot, a single checkpoint jointly serves as a play policy and expected-outcome predictor, enabling real-time identification and collection of model--reality mismatches for future self-improvement.
\end{abstract}

\IEEEpeerreviewmaketitle
An embodied agent must answer two complementary questions: what will happen after executing an action, and what action will achieve a given task? World models address the former by predicting action-conditioned futures for imagined rollouts, planning, and control \cite{finn2016unsupervised,finn2017deep,hafner2019learning,hafner2025mastering,khorrambakht2025worldplanner}; policies and inverse-dynamics models address the latter by generating feasible actions. Since these capabilities are complementary views of the same state--action trajectory distribution, jointly modeling them promises a unified model that can act, predict, and evaluate action consequences.

Recent work has pursued this objective through joint video--action generative models. GR-1 jointly denoises future images and actions \cite{wu2024unleashing,guo2024prediction}, while UVA learns a shared representation with decoupled diffusion heads for policy learning, forward and inverse dynamics, and video prediction \cite{li2025unified}. Most closely related to our setting, UWM couples video and action diffusion in a shared transformer with independently controlled diffusion timesteps \cite{zhu2025unified}. Varying the state--action noise configuration allows a single UWM model to operate as a policy, world model, inverse-dynamics model, or video generator. More recently, pretrained video generators have also shown promise as scalable world--action models \cite{bi2026motus,ye2026world}.

However, joint modeling and flexible inference do not by themselves guarantee balanced learning, stable long-horizon generation, or real-time open-loop simulation. Most existing approaches either rely on fixed-horizon, computationally heavy video backbones that are poorly suited to streaming and infinite-horizon generation, or expose such capabilities only as a simulator \cite{hafner2025trainingagentsinsidescalable} or a policy \cite{ye2026world}.

We introduce FLEX-WAM, a Flexible and EfficientNet Block-Causal World--Action Model, together with a training procedure for unified simulation and policy inference. FLEX-WAM supports variable-length context histories, variable-length non-causal prediction horizons, and infinite autoregressive rollouts generated block by block or frame by frame. Its block-causal, KV-cacheable architecture enables stable rollouts over thousands of real-time steps while supporting efficient real-time generation.

A central challenge in training a unified world--action model is preserving simulation capability (action responsiveness) while learning history-driven state--action prediction vector fields in policy and video-generation modes. We address this challenge with two training-time components. First, we numerically estimate and balance the gradient contributions of independent state and action flow-matching losses across the state--action noise grid. Second, we introduce \emph{Forward-Dynamics (FD) elasticity}, an efficient proxy for world-model action responsiveness. We monitor its degradation throughout training and use it to adaptively bias sampling toward world-model mode.

Because a world--action model must first learn action-conditioned dynamics to serve as both a simulator and generator, our training procedure begins in pure world-model mode and gradually transitions to joint WAM training. During this transition, FD elasticity regulates the sampling schedule to preserve world-modeling capability.

We evaluate FLEX-WAM's efficiency, long-horizon stability, and unified multi-mode capabilities by deploying it jointly as a simulator and action proposer within Monte Carlo Tree Search (MCTS). Planning entirely in FLEX-WAM's imagination, the resulting system solves long-horizon, multi-step puzzle tasks from OGBench \cite{park2025ogbench} as well as real-world PushT motion-planning tasks. We further present preliminary bimanual-robot results in which a single FLEX-WAM checkpoint jointly serves as a policy and expected-outcome predictor. This unified deployment enables real-time detection and automatic collection of counterfactual examples, providing a path toward improving learned world models through real-world interaction.

Our contributions are summarized as follows:
\begin{itemize}
    \item We introduce FLEX-WAM, an open-source\footnote{Code and checkpoints are omitted for anonymous review and will be released upon acceptance} block-causal world--action model that supports frame-by-frame and block-by-block autoregressive prediction, variable context horizons, and KV-cacheable rollouts stable over thousands of real-time steps.
    
    \item We introduce a gradient-calibrated, feedback-regulated two-stage training procedure that measures and preserves the action responsiveness required for temporally dense simulation.
    
    \item We evaluate the proposed architecture on simulated and real-world datasets, demonstrating competitive prediction quality and rollout efficiency relative to representative state-of-the-art world--action models.
    
    \item We show that FLEX-WAM can solve long-horizon planning tasks entirely in imagination when deployed jointly as an action proposer and simulator within MCTS.
    
    \item We demonstrate the potential of unified world--action models to identify and collect counterfactual examples for self-improvement during real-world deployment.
\end{itemize}

\section{Method}

\subsection{WAM as a Joint Flow Matching Model}
\label{sec:joint}

We define a WAM as a joint generative model of state ($z_t$) and action ($a_t$) sequences:
\begin{equation}
  p_\theta\!\left(z_{t:t+H},\, a_{t:t+H} \right),
  \label{eq:joint}
\end{equation}
%
We merge ideas from~\cite{hafner2025trainingagentsinsidescalable, chen2024diffusion} and model this distribution as a flow matching model with per-modality and per-timestep independent state and action signal levels $\tau_{s_t}$ and $\tau_{a_t}$. The per-modality independence allows diverse training and inference modes including world model ($\tau_{a}=1, \tau_s \in [0,1]$), inverse dynamics ($\tau_{a}\in [0,1], \tau_s = 1$), video model ($\tau_{a}=0, \tau_s \in [0,1]$) and joint state-action generative model  ($\tau_{a}\in [0,1], \ \tau_s \in [0,1]$). On the other hand the per-frame noise control allows any subset of the model input tokens to serve as clean context or prediction horizon and improves the model long-horizon rollout stability by exposing it to various context perturbations that effectively reduce the distribution mismatch between a context filled with the model's own predictions during inference and ground-truth during training time \cite{chen2024diffusion}.  

\begin{figure*}[t]
  \centering
  \includegraphics[width=0.9\textwidth]{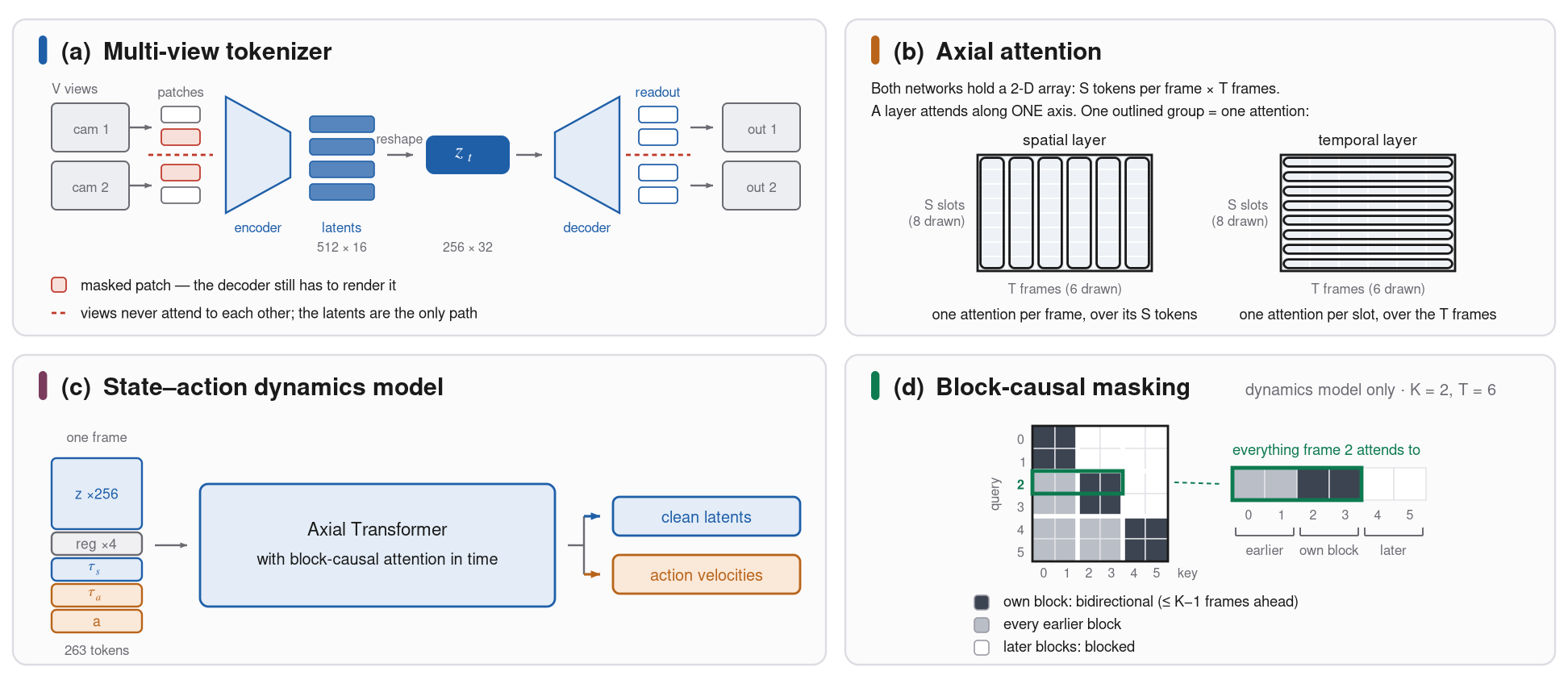}
  \caption{\textbf{WAM.} (a)~The tokenizer encodes every camera view into one
  shared set of latents per frame, and is trained to reconstruct masked patches;
  views never attend to each other, so the latents are the only path between them.
  (b)~Both networks hold their tokens on a frame $\times$ token grid and attend
  along one axis per layer. (c)~The dynamics model denoises one $263$-token set per
  frame --- latents, registers, the two noise levels, and the action --- into clean
  latents and action velocities. (d)~Its attention in time is block-causal: a frame
  attends to its own block in both directions and to every earlier block, never to
  a later one.}
  \label{fig:overview}
\end{figure*}
\subsection{FLEX-WAM: Architecture}
Our core axial attention efficient transformer architecture and multi-modal tokenizer are based on the DreamerV4 world model \cite{hafner2025trainingagentsinsidescalable} (Fig.~\ref{fig:overview}). However, we extend this framework both architecturally and training-wise both necessitated by the joint state-action modeling.  

\subsubsection{Block-Causal Axial Transformer}
\label{sec:backbone}

We build our tokenizer and dynamics model on top of block-causal axial transfomer blocks described in this section. Stacking per-timestep token sets $S$ along time $T$ and fusing them spatiotemporally is subject to $O((TS)^2)$ compute complexity due to attention operations. Instead, we adopt axial attention where each layer only fuses information spatially or temporally and is subject to lower computational complexity of $O(TS^2 + ST^2)$ \cite{ho2019axialattentionmultidimensionaltransformers}. We adopt RoPE\cite{su2024roformer} as positional embedding in each layer and interleave the architecture with AdaLN layers in case text/task conditioning is required (in this paper we do not use text/class conditioning because text and action are competing information regarding the future of a rollout and impair the model's capability as a simulator). 

Instead of training the model fully causally as in~\cite{hafner2025trainingagentsinsidescalable} (reduced quality in predicting long horizons), fully bi-directionally~\cite{zhu2025unified} (not KV-cacheable), or with partitioned causal context and bi-directional horizon~\cite{agarwal2026cosmos} (committing to a fixed horizon length set in training time), we propose variable block size block causal attention as a randomized knob during training and configurable setting during inference. Specifically, frames are partitioned into contiguous blocks of $B$ frames where frame $q$ may attend to frame $k$ iff $b(k) \le b(q)$, where $b(\cdot)$ is the block index: attention is bidirectional within a block and strictly causal across blocks (Fig~\ref{fig:overview}.d). During training, we draw $K$ once per forward pass, from $\{1,2,4,8,16\}$, shared by every sequence in the batch. As a result, block size is a deployment-time choice that allows our model to operate either completely causally ($B=1$) or block-by-block ($B>1$) when a predictive horizon larger than one is needed.

\subsubsection{Multi-View Tokenizer}
\label{sec:tokenizer}

As formulated in \cite{hafner2025trainingagentsinsidescalable}, our tokenizer is a masked auto-encoder~\cite{he2021maskedautoencodersscalablevision}, trained to reconstruct spatially and temporally randomly masked modalities. The encoder attention mask is set such that, each modality can only attend to itself while the latent tokens can attend to the modalities and themselves. On the other hand in the decoder each modality can attend to itself and the latent while the latent can only attend to itself. This ensures that during encoding, cross modality information has passed the latents and in the decoding they are decoded purely from latents and not affected by each other. In this paper, encoder and decoder are both fully causal in time ($B=1$) and block sizes larger than one is unnecessary. 

\subsubsection{Dynamics Model (WAM)}
\label{sec:dynamics}
As mentioned in Sec.~\ref{sec:joint}, we implement the WAM dynamics as a flow matching model with the following tokens per each timestep:
\begin{equation}
  \big[\;\underbrace{z_t^{1:256}}_{\text{state}} \;\big|\; \underbrace{r^{1:4}}_{\text{registers}} \;\big|\; \underbrace{\tau_{s_t} \;\big|\; \tau_{a_t}}_{\text{noise levels}} \;\big|\; \underbrace{a_t}_{\text{action}}\;\big],
  \label{eq:tokens}
\end{equation}
thus $S\!=\!263$ tokens in our implementation. The two signal levels --- $\tau_s$ for the states,
$\tau_a$ for the actions are discrete inputs with 128 possible levels (to allow future extension to short-cut models) and are sampled independently from a square grid of $128 \times 128$. The register tokens $r^{1:4}$ are placeholders for the model to carry forward information from the past if needed and operate like a scratchpad areas/short-term memory slots for the model~\cite{hafner2025trainingagentsinsidescalable}.
\section{Training}
\subsubsection{Blockwise Diffusion Forcing}
\label{sec:bdf}
A naive adoption of diffusion forcing in our setting would sample an independent noise level per modality and timestep incurring $128\times128\times T$ possible settings. Such large sampling set implies that homogeneous coverage would necessitate millions of training steps. To mitigate this issue and better represent the actual inference time applications where full blocks are denoised at a time, we sample per-modality noise at temporally identical noise levels within each block. Combined with random selection of block sizes,  our sampling will asymptotically cover per-frame diffusion forcing given sufficient iterations. Finally, for each block we use x-prediction to train the state velocity field in order to assign more model capacity to high-noise portions of the flow path and v-prediction for actions in order to focus network capacity toward cleaner portions.

\subsubsection{Action-State Gradient Normalization}
\label{sec:gradnorm}

In a joint multi-stream architecture like ours, the state and action losses $\mathcal{L}_{obs}, \mathcal{L}_{act}$,
compete for one set of shared network parameters. Added together at fixed per-modality weight, the gradient force from each channel is not guaranteed to reach the shared trunk equally and across  flow-matching noise grid. 

We weight the two
flow-matching losses, $\mathcal{L}_{\mathrm{obs}}$ by $w_{\mathrm{obs}}$ and
$\mathcal{L}_{\mathrm{act}}$ by $w_{\mathrm{act}}$,
\begin{equation}
  w_m \;=\; \frac{1}{Z}\cdot\frac{\Phi_m \, \mathrm{ramp}_m}{S_m \sqrt{\hat{L}_m}} ,
  \label{eq:wm}
\end{equation}
built from two factors set by the block's noise levels, $\Phi_m$ and
$\mathrm{ramp}_m$; a measured gradient scale $S_m\sqrt{\hat{L}_m}$; and one rescale
$Z$ common to both losses. 
\paragraph{$\Phi$ and $\mathrm{ramp}_m$} $\Phi$ splits a unit weight across two losses using the \emph{other} quantity's cleanness,
\begin{equation}
  \Phi_{\mathrm{act}} = \frac{\tau_s + \varepsilon}{\tau_s + \tau_a + 2\varepsilon},
  \qquad
  \Phi_{\mathrm{obs}} = \frac{\tau_a + \varepsilon}{\tau_s + \tau_a + 2\varepsilon},
  \label{eq:phi}
\end{equation}
summing to $1$ ($\varepsilon\!=\!10^{-3}$), so world-model weight is high when action is clean and the state is not, and vice versa. The
ramp $\mathrm{ramp}_m = \beta_m + (1-\beta_m)\tau_m$ instead weights a loss by its
\emph{own} cleanness, low noise counting more than high. It is the same ramp used in Dreamer~4~\cite{hafner2025trainingagentsinsidescalable}, with $\beta_{\mathrm{obs}}\!=\!0.2$.

\paragraph{Calibration} We reweight both losses by the denominator
of~\eqref{eq:wm} to equal gradient magnitude, and to modulate their relative
importance solely with $\Phi$ and the ramp. $\hat{L}_m$ is an EMA of the unweighted loss $m$, tracked
over the noise grid. $S_m$ is the norm of the trunk gradient of $\mathcal{L}_m$
alone, re-measured every $500$ training steps by backpropagating each loss term on
its own.

\paragraph{Action Sensitivity Regulation}
\label{sec:fd-feedback}
In a temporally dense WAM architecture such as ours, consecutive frames are highly correlated, creating an imbalance between the conditioned and unconditioned vector fields in world-model mode, $v(z_{t+1}|a_t,z_{t-C:t})$ and $v(z_{t+1}|\phi,z_{t-C:t})$.  

To ensure the world model capability is preserved, we propose a dual stage training protocol: we first train the model in world model mode only ($\tau_a=1, \tau_s \in [0,1]$) while estimating the normalized sensitivity of the latents to action perturbations around a subset of evaluation windows using a Hutchinson estimator. Calling this the Forward Dynamics (FD) elasticity sensor, we empirically observe that this sensitivity metric reaches a smaller peak compared to a purely WM-mode training under uniform noise sampling for state and action (WAM training protocol). Thus, in the second stage after the metric converges in WM-only mode, we define a closed-loop feedback regulator with a setpoint chosen at 90 percent of WM-only sensitivity and the control input being the sampling bias of WM mode during training (Sec.~\ref{sec:exp-fd-sensor}).  
\begin{table*}[t]
    \centering
    \caption{Experimental domains and their roles in our evaluation.}
    \label{tab:experimental_setup}
    \footnotesize
    \setlength{\tabcolsep}{5pt}
    \renewcommand{\arraystretch}{1.12}
    \begin{tabularx}{\textwidth}{
        @{}
        l
        l
        >{\raggedright\arraybackslash}X
        c
        >{\raggedright\arraybackslash}X
        @{}
    }
        \toprule
        \textbf{Domain}
        & \textbf{Dataset / Platform}
        & \textbf{Evaluation Purpose}
        & \textbf{Model Size}
        & \textbf{Primary Evidence} \\
        \midrule
        Simulation
        & LIBERO
        & Generation quality and inference latency
        & 690M
        & SSIM, PSNR, and latency \\

        Real-world play
        & PushT
        & Action sensitivity and planning
        & 400M
        & DINO response and imagined planning \\

        Puzzle, Cube, Scene
        & OGBench
        & Long-horizon planning
        & 512M
        & Imagined task success \\

        Humanoid
        & Unitree G1
        & Long-horizon rollout
        & 690M
        & Qualitative rollout stability \\

        Deployment
        & OpenArm
        & Model-reality mismatch detection
        & 690M
        & Closed-loop case study \\
        \bottomrule
    \end{tabularx}
\end{table*}

\section{Results}

\subsection{Experimental Setup}
We evaluate FLEX-WAM across five domains chosen to isolate core capabilities: generation quality and inference
efficiency on LIBERO, action sensitivity and planning on
PushT, long-horizon planning on OGBench, long-rollout
generation on Unitree G1 data, and closed-loop deployment on
OpenArm. As action-faithful world modeling requires
exposure to outcomes beyond predominantly successful
demonstrations, our real-world experiments primarily use
teleoperated play data containing diverse and often
suboptimal interactions. Table~\ref{tab:experimental_setup} summarizes each domain's role.


For PushT, we use a four-hour real-world play
dataset for action-sensitivity evaluation and planning \cite{khorrambakht2025worldplanner}. We
additionally use the OGBench $4\times4$ visual-puzzle task
for long-horizon planning \cite{park2025ogbench}.

We validate the scalability of our model in learning complex embodiments and tasks, by training on the play dataset from \cite{dmo-rl} of a Unitree G1 humanoid robot with a head-mounted camera and BrainCoV2 dexterous hands manipulating a bag with flexible handle which forces the model to capture the intricate complexities of the fingers and handle interaction only from a ego-centric camera \ref{fig:long-rollout}. Finally, we show preliminary results on the ability of FLEX-WAM in self-discovering corner-cases where the world model is weak, through a short play dataset and an OpenArmV1 bi-manual station manipulating an articulated drawer toolbox (Fig.~\ref{fig:counterfactual}).

Based on task and embodiment complexity, we train our model at three scales: (690M for G1, LIBERO, and OpenArm and 512M for OGBench, and 400M for PushT) under Distributed Data Parallel (DDP) training on 8 NVIDIA H100 GPUs each for roughly 150k gradient steps (4 days). The models are all initialized randomly (no large-scale video pretraining), and with effective batch size of 64, and context length 32.
Finally, our local inference results are carried out using up to 4 NVIDIA RTX6000 Pro MaxQ cards (for real-time inference with batch size of 1 a single card is sufficient). We adopt bfloat16 mixed precision and torch.compile optimizations to accelerate training and inference. 


\subsection{Design Choices}
In this section we empirically validate our action sensitivity FD sensor and the dual-stage training protocol in Sec.~\ref{sec:fd-feedback}, and the importance of block-causality in improving temporal consistency in horizon prediction. 

\begin{figure}[t]
    \centering

    \subfigure[]{%
        \includegraphics[width=0.45\columnwidth]{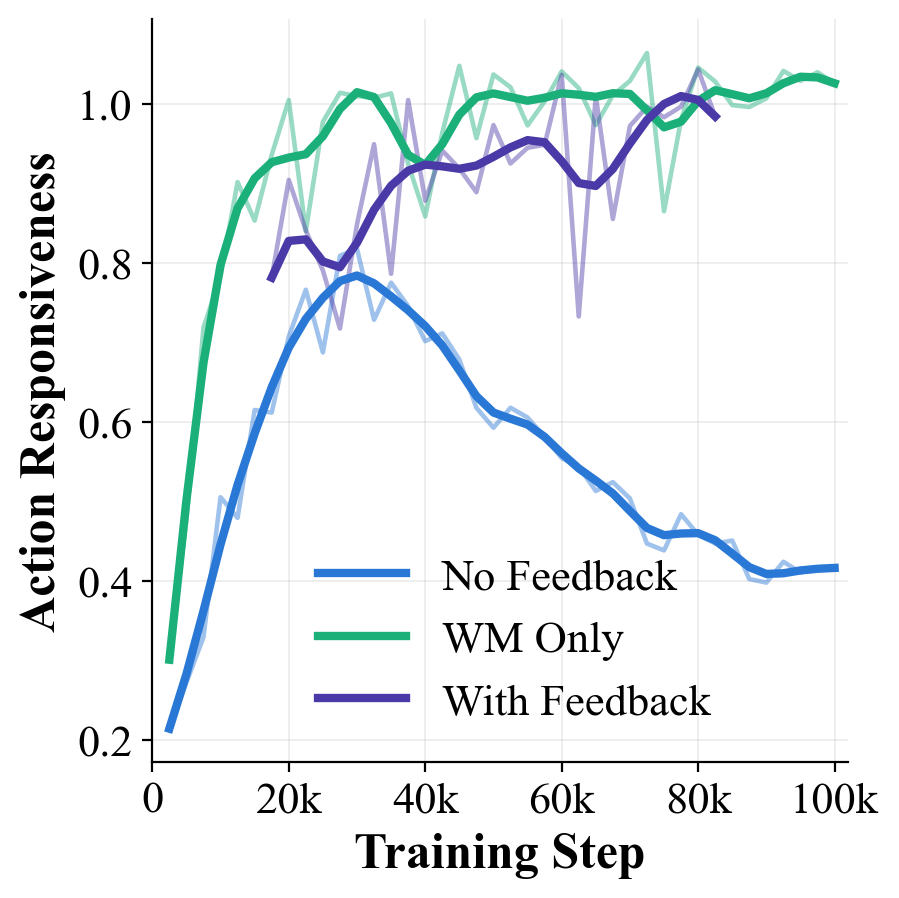}%
        \label{fig:action-response-probe}
    }\hfill
    \subfigure[]{%
        \includegraphics[width=0.45\columnwidth]{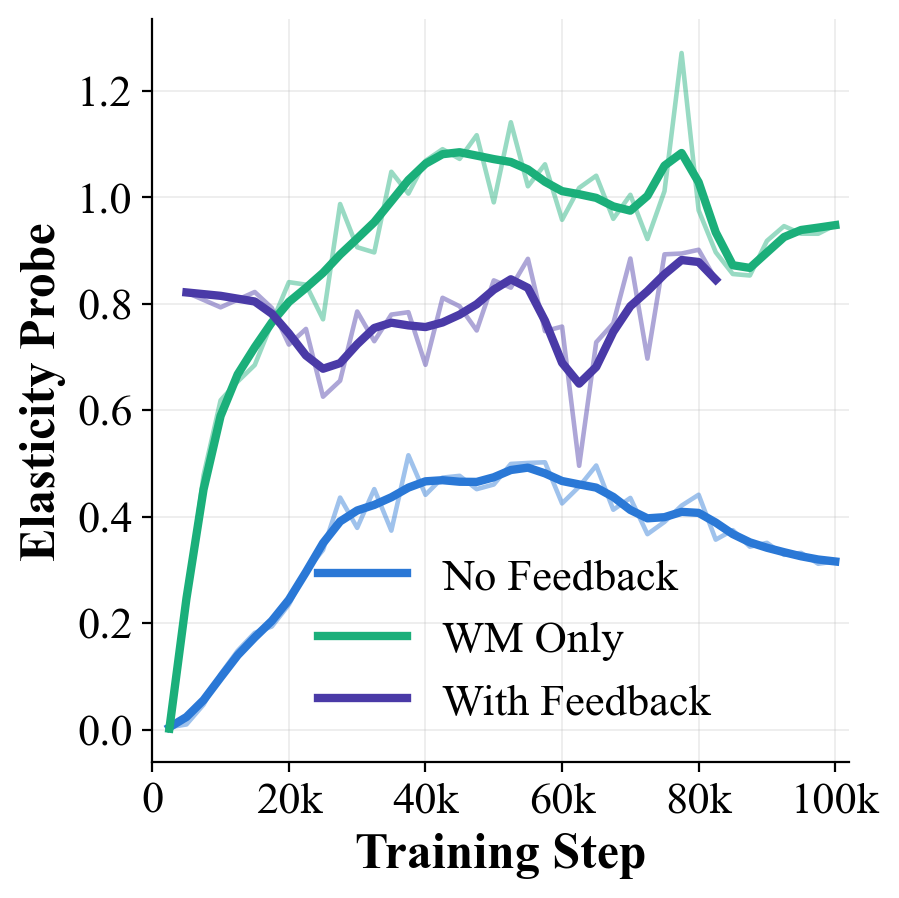}%
        \label{fig:action-response-fd}
    }
    \caption{WAM response to synthetic counterfactual stop command along a random set of PushT dataset trajectories with and without training-time feedback regulation.}
    \label{fig:action-response}
\end{figure}

\subsubsection{Dual-Stage Training Protocol and FD-Sensor Validity}
\label{sec:exp-fd-sensor}
In a temporally dense WAM architecture such as ours, consecutive frames are highly correlated, creating an imbalance between the conditioned and unconditioned vector fields in world-model mode, $v(z_{t+1}|a_t,z_{t-C:t})$ and $v(z_{t+1}|\phi,z_{t-C:t})$. Classifier-free guidance (CFG) could rebalance their relative strengths at runtime, but would require two GPUs to maintain minimal latency.

As described in Sec.~\ref{sec:fd-feedback}, we instead propose training-time closed-loop regulation based on FD elasticity, an efficient proxy for the model's responsiveness to actions. We test whether this proxy correlates positively with world-model responsiveness using a counterfactual probe. Because PushT is quasi-static, we construct a synthetic case in which the WM follows actions from dataset trajectories and then stops when the action is set to null: zero for relative actions, or the last action for absolute actions. The expected ground truth is the final state before the pause. We measure the distance from the generated image to this state using DINOv2 features, and interpret zero distance as perfect action response.

Fig.~\ref{fig:action-response-fd} shows FD sensitivity throughout training across three variants, normalized by the maximum sensitivity reached by the WAM trained purely in world-model mode ($\tau_a=1$). Fig.~\ref{fig:action-response-probe} shows the corresponding capability measured by the DINO probe, normalized by the best response of a WM-only model. This clearly shows the advantage afforded by our feedback mechanism.

WM-only model is trained in world-model mode; its capability rises during training and converges to a stable baseline. No Feedback denotes our WAM trained normally without capability measurement or regulation. Its capability rises to a peak below the WM-only baseline and then declines. With Feedback is first trained in WM-only mode until convergence, after which a closed-loop WM-only sampling bias against other modes is activated and regulated using the training-time FD elasticity (WM capability) measurement. This regulates WM action sensitivity around a setpoint around the peak achieved in WM-only pre-training while gradually permitting non-WM modes. In our experiment, the WM-mode sampling bias starts near one and converges to approximately 0.2 following a nontrivial schedule that would be difficult to hand-tune as a fixed parameter if not controlled through feedback.

\subsubsection{Block-Causality vs Fully Causal Horizon}
\begin{figure}[t]
    \begin{center}
        \includegraphics[width=0.95\linewidth]{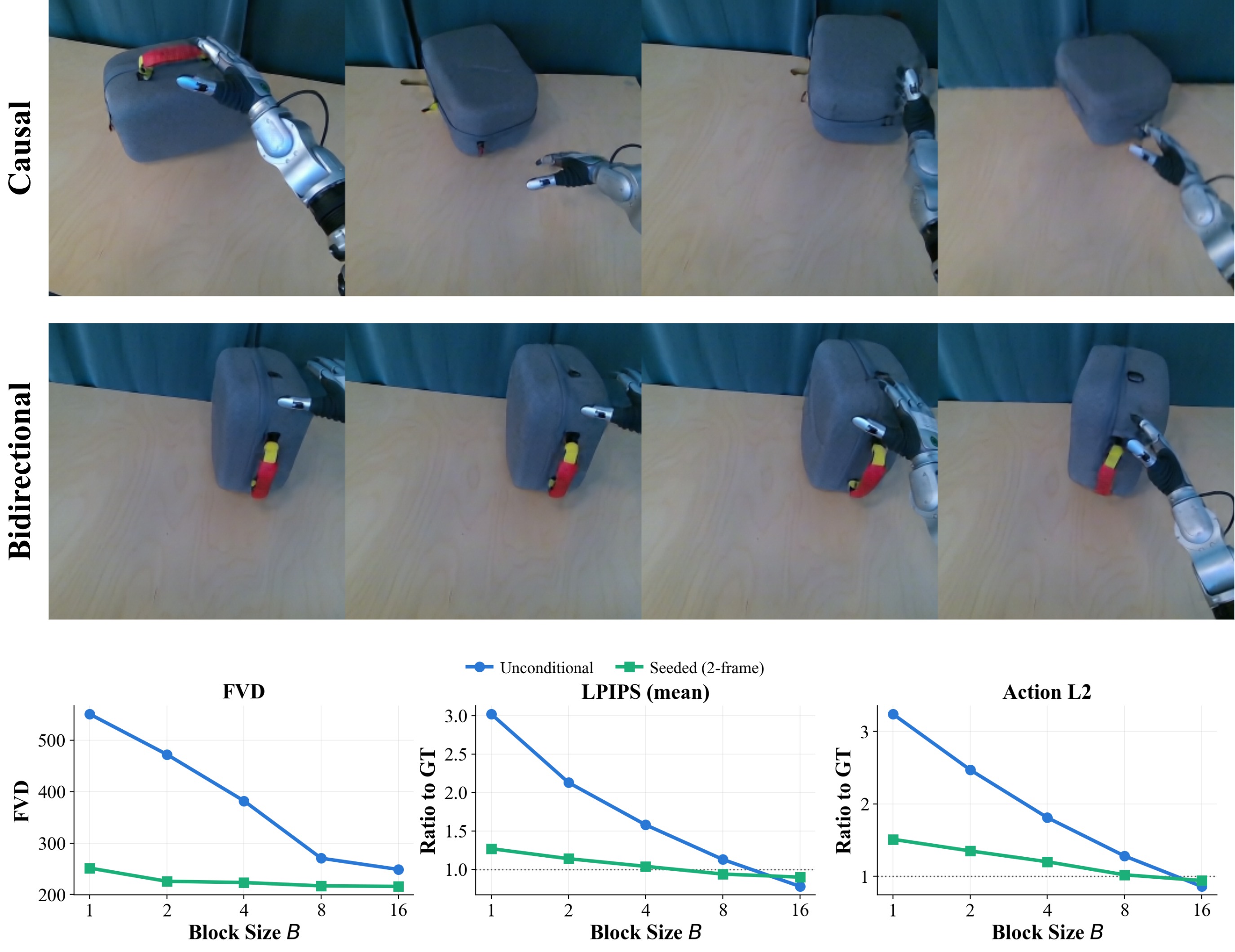}
        \caption{Bi-directional attention along the horizon facilitates temporal consistency and smoothness by allowing past frames to attend to the future ones during denoise process. Snapshots show decoded frames from FLEX-WAM with no seed history frames under causal and fully bidirectional attention showing broken temporal consistency in fully causal mode. Plots quantify this by measuring cross frame visual and action distances and dataset-model FVD distance.}
        \label{fig:bidirectional-attention}
    \end{center}
\end{figure}

In model-predictive control and accurate block-by-block rollout stability, a world action model should be capable of rolling out from the current moment up to a fixed horizon without breaking the temporal consistency between the frames. Flex-WAM being trained on variable blocks of bidirectional attention allows the frames within a chosen horizon during rollout attend to each other in both temporal directions. Here we show how bi-directionality indeed facilitate temporal consistency by denoising a fixed horizon of 16 frames in parallel while sweeping the block attention width from 1 (fully causal) to 16 (fully bidirectional). We conduct this experiment on a FLEX-WAM trained on the G1 dataset and show the qualitative and quantitative results in Fig.~\ref{fig:bidirectional-attention}. The FVD~\cite{unterthiner2019fvd} plot measures the semantic video distance between 256 seeded and unconditional (no history of the past frames) generated sequences vs 256 ground truth clips and its lower value indicates a higher spatiotemporal similarity to the ground-truth. We also measure the temporal smoothness by computing the LPIPS and L2 distances between consecutive frames and actions respectively, and normalized against their average value in training dataset. A lower number for these two metrics indicate closer similarity in terms of motion and action smoothness to the ground truth. As shown both by the qualitative snapshots and the plots, bidirectionality is positively correlated with the temporal smoothness and consistency with the ground truth. We also interestingly note that when the model is seeded with past frames, the pressure to attend to the future is much lower. We hypothesize that the model predictive uncertainty (e.g. multi-modal futures) is positively correlated with stronger reliance on non-causal temporal attention activities and leave it as a promising future research direction.  

\subsection{Latency and Long-Horizon Generation Stability}
We now compare our FLEX-WAM architecture with other joint video-action generative models to test its relative inference speed and long-horizon generation quality.
\begin{figure}[!t]
    \centering
    \subfigure[]{%
        \includegraphics[width=0.45\columnwidth]{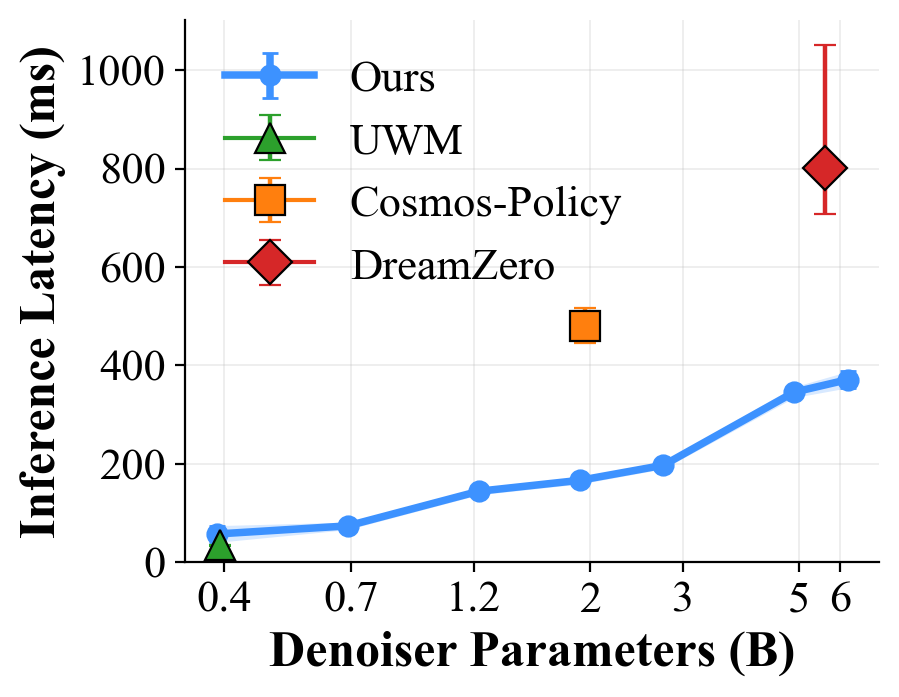}%
        \label{fig:parameter-size-scaling}
    }\hfill
    \subfigure[]{%
        \includegraphics[width=0.45\columnwidth]{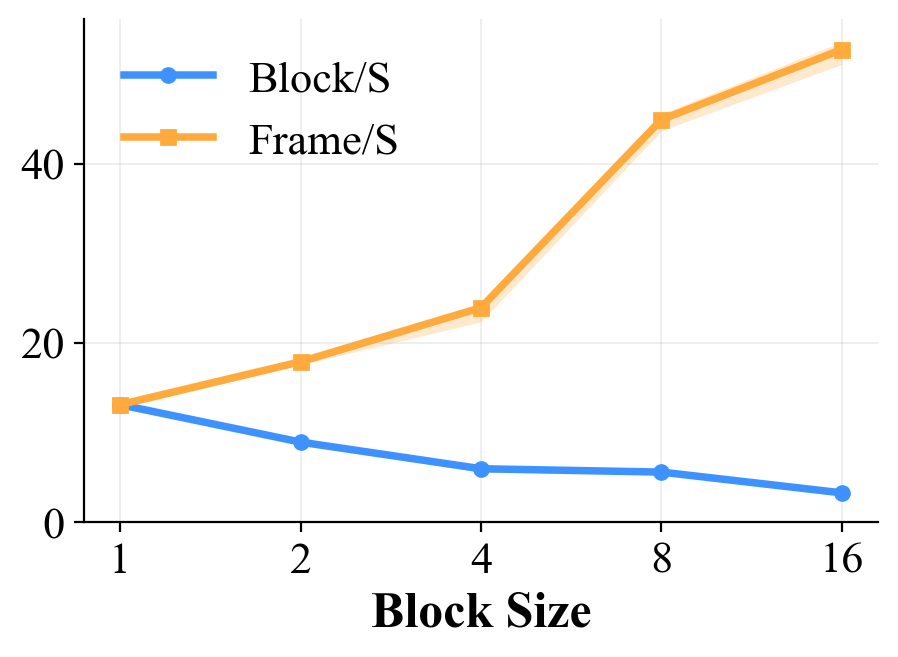}%
        \label{fig:block-size-scaling}
    }
    \caption{Inference-time scaling of FLEX-WAM.
(a) Per-step latency as a function of model size under fully
causal inference. (b) Block-generation rate and effective
frame throughput as the inference block size varies. All
block-size results use the same checkpoint; increasing $B$
trades causal latency for greater frame-level throughput.}
    \label{fig:inference-time-scaling}
\end{figure}

\begin{table}[t]
    \centering
    \caption{Inference time (latency) of our model vs.\ baselines.}
    \label{tab:baselines-timing}

    \begin{threeparttable}
        \resizebox{0.98\linewidth}{!}{%
            \begin{tabular}{lccc}
                \toprule
                & Policy & WM & Joint \\
                \midrule
                Cosmos Policy (2B) & $330.5{\pm}3.9$ & $47.0{\pm}1.0^{\dagger}$ & $480.6{\pm}7.1$ \\
                DreamZero (5B) & $801.4{\pm}102.0$ & $-$ & $801.4{\pm}102.0$ \\
                UWM (393M) & $34.1{\pm}0.2$ & $34.4{\pm}0.4$ & $34.6{\pm}0.1$ \\
                Ours (690M) & $71.5{\pm}1.0$ & $72.7{\pm}0.3$ & $71.5{\pm}1.0$ \\
                \bottomrule
            \end{tabular}
        }

        \begin{tablenotes}[flushleft]
            \footnotesize
            \item[$\dagger$] The reduced inference latency is due to the
            official implementation using a single denoising step in WM mode.
        \end{tablenotes}
    \end{threeparttable}
\end{table}

\subsubsection{Baselines}

We compare FLEX-WAM with representative joint video-action generative models. 
Specifically, UWM~\cite{zhu2025unified} has a closest formulation to our independent flow-matching objective,
while Cosmos-Policy and DreamZero build on large pretrained video-generation backbones.
To ensure fair comparison, we adopt released pre-trained checkpoints for the LIBERO benchmark (DreamZero did not have such checkpoints so we do not benchmark generation quality) and train a multi-view FLEX-WAM on the same dataset.

\subsubsection{Inference Latency}
Deployment as a policy or interactive simulator requires low-latency observation-action generation.
We evaluate FLEX-WAM in fully causal mode ($B=1$), using a 32-step KV
cache and 8 denoising steps.
Table~\ref{tab:baselines-timing} reports per-step latency native inference configuration of each
method. 
Fig.~\ref{fig:parameter-size-scaling} shows how it grows with model size. We run each baseline under the official inference setting, most importantly the number of denoising steps.
FLEX-WAM achieves competitive inference latency under
comparable model scales and denoising settings, while
additionally supporting adjustable blockwise inference.
Furthermore, Fig.~\ref{fig:block-size-scaling} shows the benefit of flexible block-size in our block-causal architecture. 
Specifically, we can maintain real-time low-latency inference speeds by choosing $B = 1$ or trade it off gradually with inference throughput by increasing the horizon length (e.g. needed in WM efficient inference in evaluating policies with action-chunk output).
Block size therefore serves as a deployment-time control over the latency-throughput trade-off without requiring
retraining.


\begin{figure*}[t]
    \centering

    \subfigure[]{%
        \includegraphics[width=0.92\textwidth]{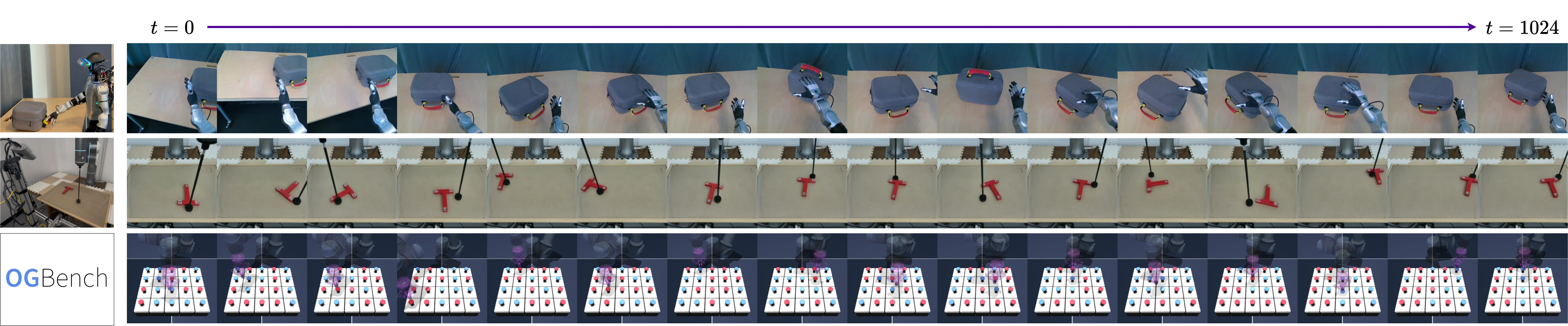}%
        \label{fig:long-rollout}
    }\hfill
    \subfigure[]{%
        \includegraphics[width=0.92\textwidth]{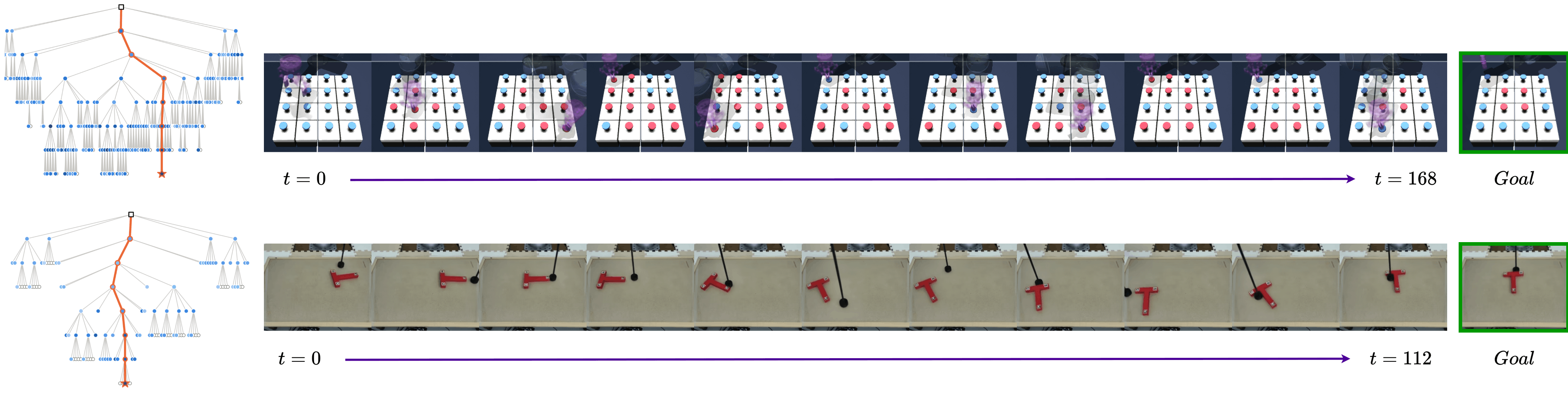}%
        \label{fig:pusht-puzzle-plans}
    }
    \caption{(a) FLEX-WAM's KV-cached block-causal architecture and blockwise diffusion forcing training enables very long stable autoregressive generation and simulation for thousands of frames. Above shows sample 1024-step autoregressive rollouts of FLEX-WAM trained on two teleoperated real-world datasets (human play) and a simulated puzzle task with $2^{16}$ states from the OGBench~\cite{park2025ogbench} benchmark. (b) Visualization of two long-horizon plans solved entirely in FLEX-WAM imagination with MCTS. On the left, the search trees: node colours are the task rewards (in $[0,1]$, light to dark), hollow rings are children created but never simulated, highlighted path is the returned plan. On the right, the decoded imagined states along the returned plan.}
\end{figure*}

\subsubsection{Generative Quality and Rollout Stability}
Thanks to per-block diffusion forcing, random context dropout, and KV-caching, our model can run autoregressively for thousands of frames either as a simulator or a real-time joint state-action streamer. Fig.~\ref{fig:long-rollout} presents the decoded dream states of our model over 1024 autoregressive steps in joint state-action prediction mode.
Across the illustrated real-world and simulated domains, the trajectories remain visually coherent throughout the rollout without obvious catastrophic accumulation artifacts.

\begin{figure}[t]
    \begin{center}
        \includegraphics[width=\linewidth]{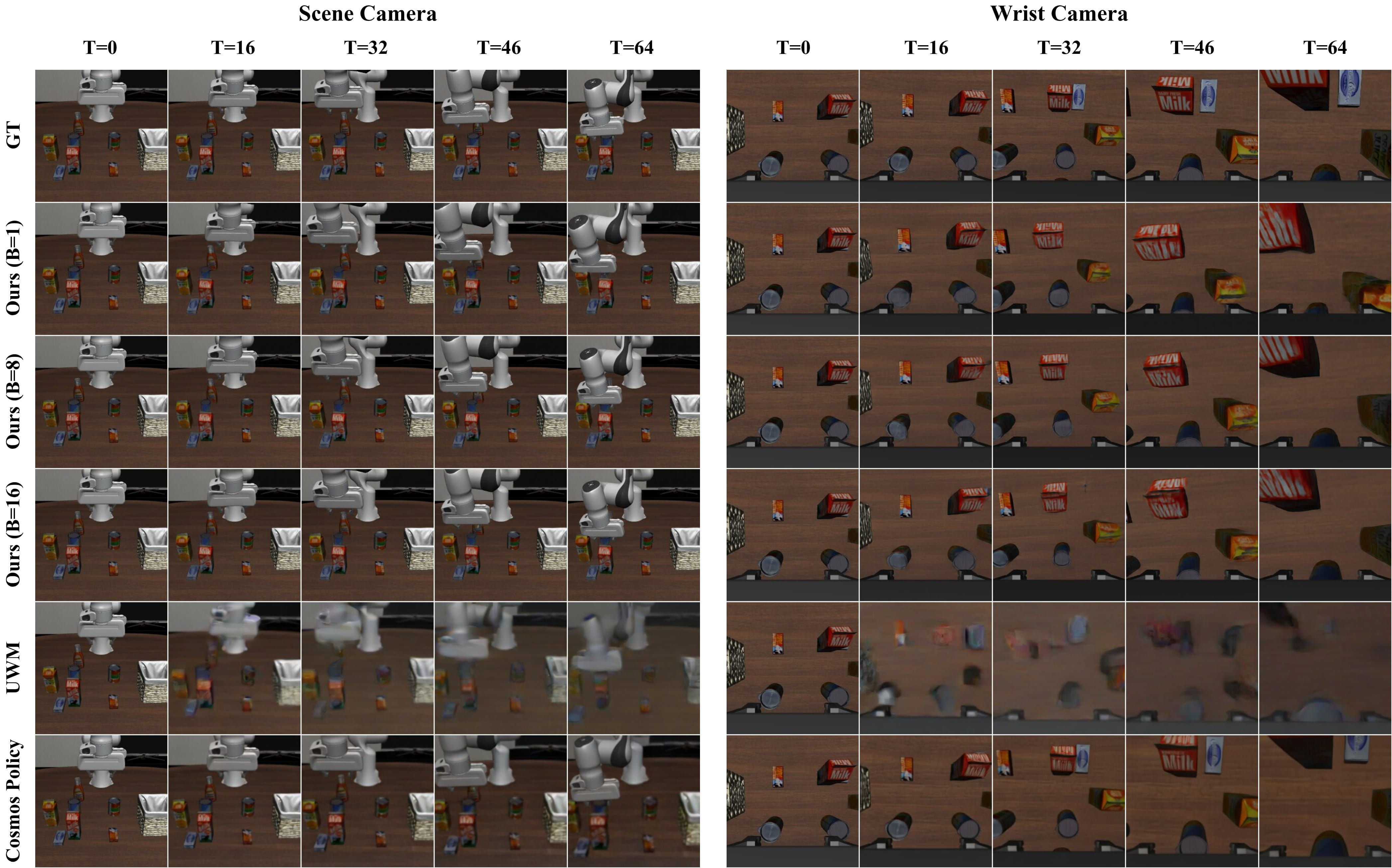}
        \caption{Autoregressive rollout in world model mode with actions provided from the evaluation dataset and decoded states visualized.}
        \label{fig:lebero-qualitative}
    \end{center}
\end{figure}

\begin{figure}[t]
    \begin{center}
        \includegraphics[width=0.95\linewidth]{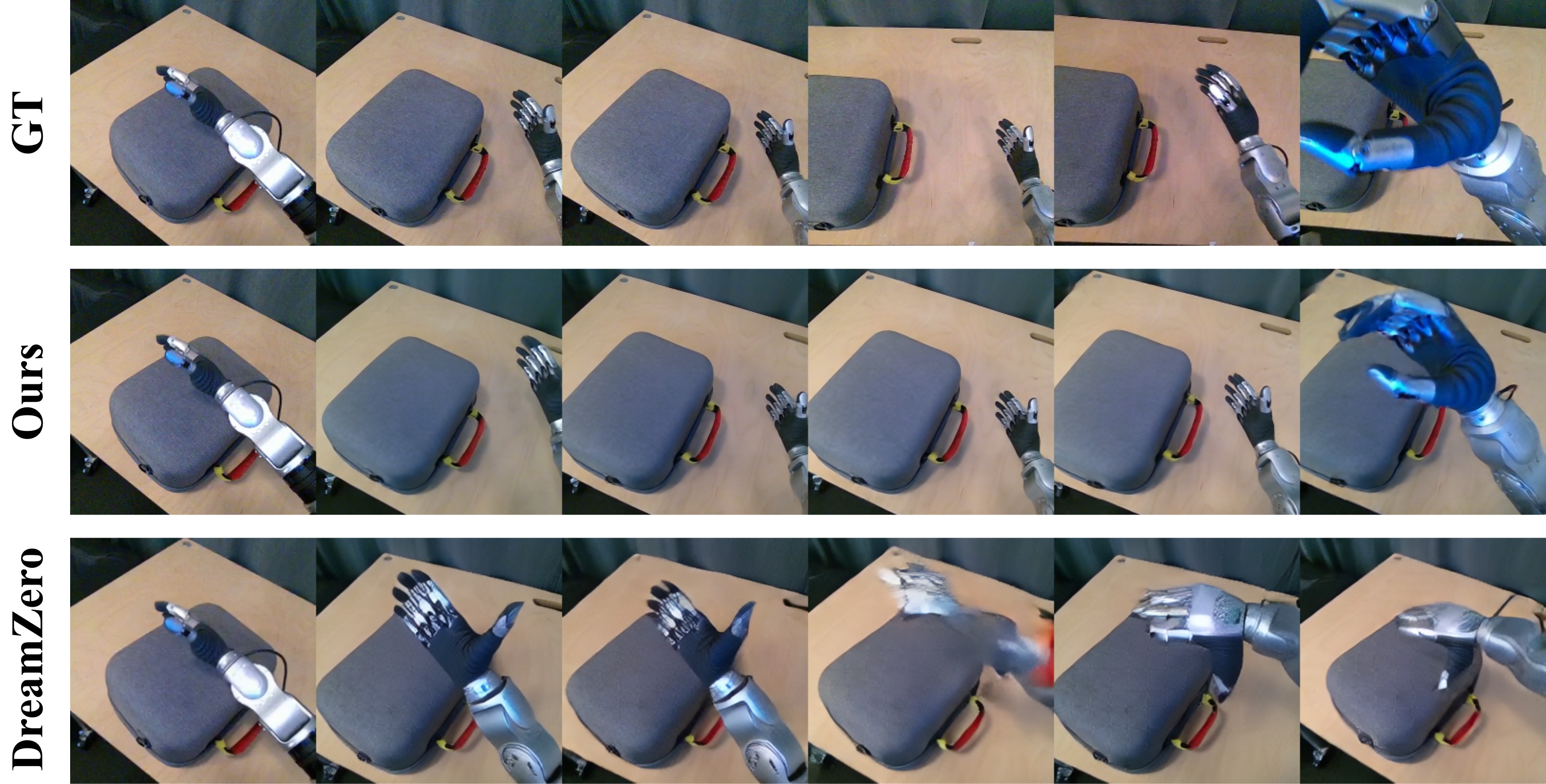}
        \caption{Open-loop rollout (4.5 seconds) of DreamZero and our model both trained under identical WM mode ($\tau_a=1$) vs ground truth.}
        \label{fig:dreamzero-g1}
    \end{center}
\end{figure}

Furthermore, Fig.~\ref{fig:lebero-qualitative} compares the generative quality of our model against baselines on the LIBERO tasks qualitatively and Table.~\ref{tab:rollout} reports SSIM and PSNR over prediction horizons from $T=16$ to $T=64$. 
To ensure comparisons are fair, we only compare in world model mode where actions ground all the models to the same future outcome. Notably both baselines predict the terminal state given the current observation and a fixed-length action chunk with no KV-caching feature. Thus, we compute future states by feeding their context with the current prediction and jumping a chunk length at a time. As shown, our model exhibits highly stable generation and quality (after decoding). Because DreamZero does not support world model mode officially, we train it on our G1 example purely under WM mode ($\tau_a=1$) and compare it to our architecture trained under identical WM-only setting. Fig.~\ref{fig:dreamzero-g1} shows the qualitative results and shows that while both are trained on the full-scale dataset, our model exhibits more ground-truth consistency and no hallucinations/object deformations. We note that DreamZero as a representative of many WAMs is defined as a joint generator of state and actions (thus WAM) and is not officially intended as a simulator.

FLEX-WAM achieves higher SSIM and PSNR than the evaluated baselines at every horizon. Among the tested inference configurations, $B=8$ provides the strongest overall generation quality, while $B=16$ remains comparable at longer horizons. 

\begin{table}[b]
\centering
\caption{Generation Quality vs Horizon}
\label{tab:rollout}
\resizebox{\linewidth}{!}{
\begin{tabular}{lcccc}
\toprule
 & \multicolumn{4}{c}{\textbf{SSIM}~($\uparrow$)~/~\textbf{PSNR}~($\uparrow$,\,dB)} \\
\cmidrule(lr){2-5}
\textbf{Baseline} & $\mathbf{T=16}$ & $\mathbf{T=32}$ & $\mathbf{T=48}$ & $\mathbf{T=64}$ \\
\midrule
UWM & 0.652/18.3 & 0.624/17.1 & 0.610/16.4 & 0.599/15.6 \\
Cosmos-Policy & 0.682/20.4 & 0.626/18.3 & 0.604/17.2 & 0.587/16.5 \\
Ours (B=1) & 0.828/23.9 & 0.797/21.4 & 0.778/20.3 & 0.764/19.9 \\
Ours (B=8) & 0.843/24.8 & 0.812/22.5 & 0.788/21.1 & 0.778/20.5 \\
Ours (B=16) & 0.828/24.1 & 0.806/22.3 & 0.787/21.1 & 0.776/20.5 \\
\bottomrule
\end{tabular}
}
\end{table}

\subsection{Long-Horizon In-Imagination Planning}
Solving long horizon tasks through planning requires a stable world model to simulate future outcomes of potential actions, and an affordance-aware action sampler to reduce the planning search space to plausible action primitives. Our FLEX-WAM highly stable long-horizon joint world-action rollout capability alongside its chunked prediction mode can serve as unified simulation and action sampler component in MCTS. Specifically, it accelerates planning through two mechanisms: 1) variable causal attention block-size allows choosing $B>1$ during sampling thus cutting the autoregressive steps by a factor of $B$ and 2) it predicts state and action jointly at the same time thus removing the need for step-by-step third-party policy rollout in feedback with our model as a simulator. In this section we demonstrate how FLEX-WAM incorporated in the MCTS algorithm from \cite{khorrambakht2025worldplanner} can both solve the demonstrated PushT problem and a challenging long-horizon $4\times4$ light-out puzzle problem from the OGBench benchmark~\cite{park2025ogbench}. 

\subsubsection{FLEX-WAM in MCTS}


Each tree node holds a latent state together with a short context of past frames (at most $3$ frames of states and actions). An edge is a rollout of FLEX-WAM in joint state-action mode, $H$ frames long, that represents a random affordance-aware primitive action when the robot is at a particular node. Due to the inference-time stochasticity ($z_0 \sim N(0, I)$) and goal-free nature of the play dataset, each inference produces a different plausible future (MCTS with a task-conditioned WAM is less challenging to solve due to the sampling strong bias towards expert actions). An edge of $H$ frames is generated in $H/B$ successive calls of $B$ frames each. In our implementation, we assume $M=5$ of such samples in each tree expansion step (parallelizable over multiple GPU). Each leaf is selected based on the UCB1 score, we run a $H_{\text{sim}}$-step simulation and scored it using a specific rule that depends on the task considered. Each new child is checked against a goal test on its terminal state. The search stops as soon as a child corresponds to a terminal state, and the plan is the path from the root to that node

\subsubsection{Experiments}
Fig.~\ref{fig:pusht-puzzle-plans} shows one plan per task, each found entirely in FLEX-WAM imagination: the search tree on the left, with the returned plan in orange, and the decoded imagined states along the plan on the right. Both tasks need long sequences of intermediate steps that bring no immediate improvement, one of the benefits of tree search over greedy action selection.

\paragraph{PushT on real robot data}
The goal is to push the T to the center of table, upright (the position of the end-effector is ignored). The reward scores how close the T's position and orientation in the decoded image are to the goal ($r \in [0,1]$). A node reaches the goal when $r \geq 0.9$. A leaf is scored by the best discounted return along its $M=5$ simulated rollouts ($\gamma=0.98$): $R = \max_{m\le M} \sum_{t=0}^{H_{\text{sim}}}\gamma^t r(z_{t}^m)$. The edges and simulated rollouts are $16$ frames long, generated in two $8$-frame blocks, with a budget of $100$ MCTS iterations.. The plan in Fig.~\ref{fig:pusht-puzzle-plans} ($7$ edges, $112$ frames) turns the T from lying on its side to upright at the centre.

\paragraph{OGBench Puzzle 4x4}
We follow the evaluation protocol of OGBench's \texttt{visual-puzzle-4x4}: $5$ tasks, each given as a goal image from a given board and robot arm configuration. Each task requires 4 to 7 button presses.
We learn a goal-conditioned value function on FLEX-WAM's tokenizer latents from the play data, with the offline RL objective of GCIVL~\cite{kostrikov2021offline}. It predicts how many steps $d$ separate a state from the goal image, and the planner uses $\gamma^{d} \in [0,1]$ as reward ($\gamma = 0.99$). Because this value already measures distance to the goal, a leaf is scored by the value of the last frame of its simulated rollouts (best of $4$): $R = \max_{m\le M}\, r(z^{m}_{H_{\text{sim}}})$. The edges and the simulated rollouts are $24$ frames long, generated in $8$-frame blocks, with a budget of $600$ MCTS iterations. The search stops when goal board configuration is reached. With this learned value, the planner solves every one of the five tasks and Table~\ref{tab:puzzle-results} shows a success rate of 80\% under the OGBench benchmark settings. A qualitative visualization of one of the solved plans is shown in Fig.~\ref{fig:pusht-puzzle-plans} where the filmstrip shows the board state after each key-press. We note that in this work the success is measured as achieved in imagination: a plan counts when its imagined final state reaches the goal. Executing the plans closed loop on the real system is left to future work.

\begin{table}[t]
\centering
\caption{Open-loop performances of our WAM model with MCTS on OGBench \texttt{visual-puzzle-4x4} evaluation set.}
\label{tab:puzzle-results}
\begin{tabular}{cccc}
\hline
\multirow{2}{*}{Task} & Solved & Median MCTS & Plan length (frames) \\
& & iterations to goal & (mean and min-max) \\
\hline
1 & 17/20 & 234 & 145 (72--192) \\
2 & 17/20 & 270 & 160 (96--216) \\
3 & 14/20 & 188 & 149 (96--168) \\
4 & 14/20 & 242 & 171 (120--240) \\
5 & 18/20 & 205 & 161 (120--240) \\
\hline
All & \textbf{80/100 (80\%)} & 228 & 157 (72--240) \\
\hline
\end{tabular}
\end{table}

\begin{figure}[t]
    \begin{center}
        \includegraphics[width=0.9\linewidth]{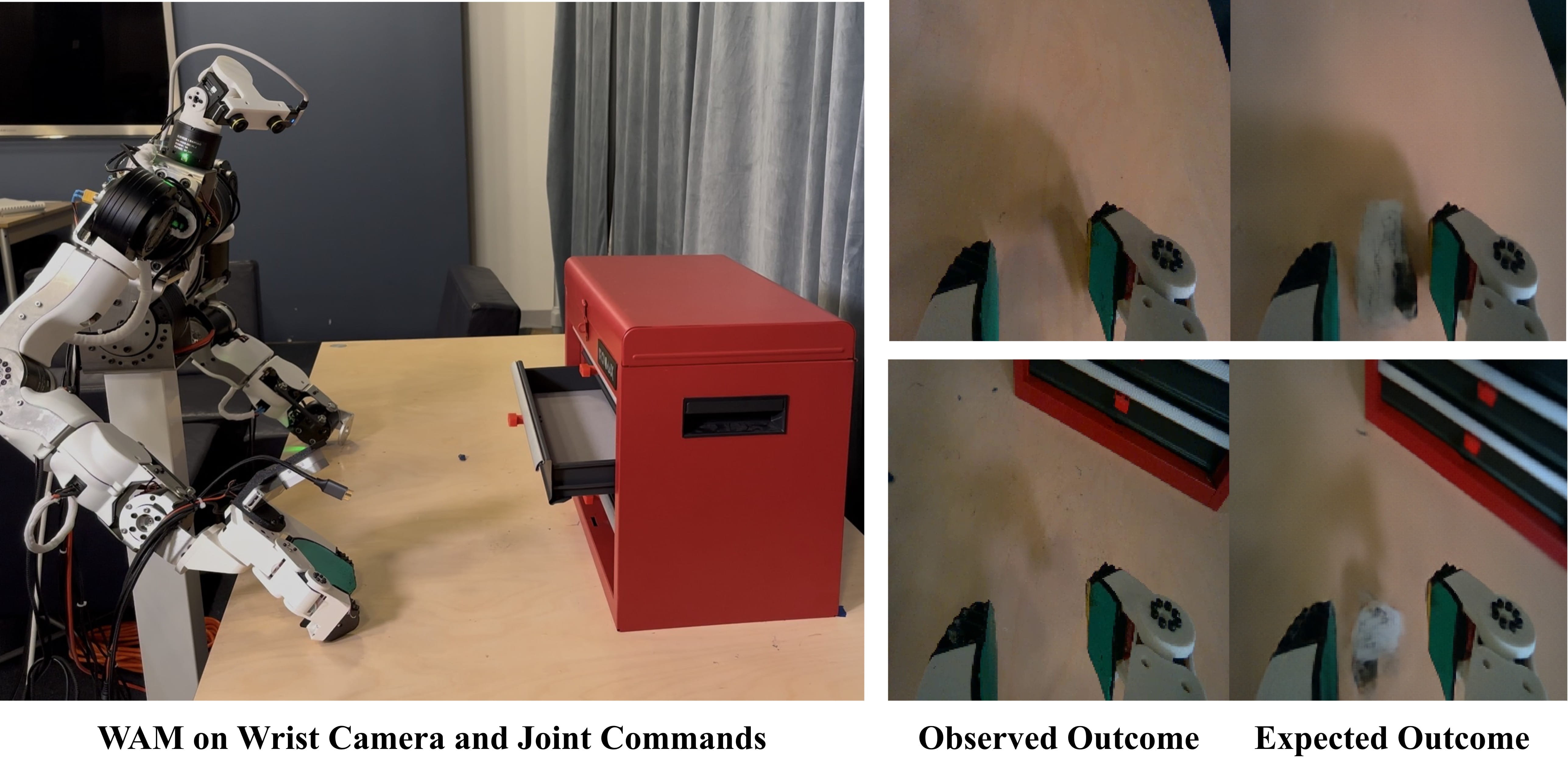}
        \caption{The joint state-action prediction capability and real-time speed of FLEX-WAM allows closed loop deployment and near-future prediction in image-space on the fly, facilitating the identification of counterfactual training samples for continual self-improvement and early warning for human intervention based training. Above shows the model trained on a small dataset hallucinating a box to pick next to the real camera feed. }
        \label{fig:counterfactual}
    \end{center}
\end{figure}

\subsection{Closed-Loop Deployment and Counterfactual Example Discovery}
Inspired by DAgger~\cite{ross2011reduction}, in this section we show how the real-time performance of our WAM deployed as a closed-loop policy (sending the real camera images into the KV-cache instead of the predicted ones) enables us to record a corrective dataset of counterfactual examples. First, we  train our WAM on an intentionally short 2-hour teleoperated play dataset using our OpenArm platform. At deployment time, the robot performs random play actions as seen in the dataset while at the same time predicting the expected outcome of each action before performing it. The difference between this prediction and real outcome then serves as an error signal for identifying and potentially up-weighing important samples in the self-recorded dataset. This has been demonstrated in Fig.~\ref{fig:counterfactual} where the WAM hallucinates a box to pick in the next timestep alongside the true observation. 


\section{Conclusion and Future Work}
\label{sec:conclusion}

We presented FLEX-WAM, a unified world-action model that
combines balanced, feedback-regulated training with adaptive
block-causal generation. Gradient calibration balances the
state and action objectives, while sensitivity regulation
preserves action-faithful world-model behavior during joint
training. Randomized block-size training and blockwise
diffusion forcing allow a single checkpoint to operate from
framewise causal prediction to higher-throughput parallel
block generation, enabling flexible and efficient
long-horizon rollouts without retraining. These capabilities
allow FLEX-WAM to provide both action proposals and
imagined outcomes for planning and to expose model-reality
mismatches during closed-loop deployment. Executing
imagined plans in the real world and using detected
mismatches for continual model improvement remain important
directions for future work.

\section{Acknowledgment}
This work was supported in part by the US National Science Foundation under Grants 2222815 and 2315396, by gifts from Wandercraft, Ultra, NVIDIA, Infinite Machines and in part by the French AI Interdisciplinary Institute ANITI. ANITI is funded by the France 2030 program under Grant Agreement number ANR-23-IACL-0002.

\bibliographystyle{IEEEtran}
\bibliography{ref2.bib}
\end{document}